\documentclass[journal,twoside,web]{ieeecolor}
\usepackage{generic}
\usepackage{cite}
\usepackage{amsmath,amssymb,amsfonts}
\usepackage{algpseudocode}
\usepackage{graphicx}
\usepackage{subcaption}
\usepackage{textcomp}
\usepackage{multirow, booktabs}
\usepackage[ruled]{algorithm2e}
\usepackage{setspace}

\def\BibTeX{{\rm B\kern-.05em{\sc i\kern-.025em b}\kern-.08em
    T\kern-.1667em\lower.7ex\hbox{E}\kern-.125emX}}
\begin{document}
\title{Computer-Aided Design Generation by Cascaded Discrete Diffusion Model}
\author{Honghu Pan, Xiaoling Luo, Yongyong Chen, \textit{Member, IEEE}, Zhenyu He, \textit{Senior member, IEEE}, Pengyang Wang
	\thanks{H. Pan is with School of Artificial Intelligence and Robotics, Hunan University. (Email: honghupan@hnu.edu.cn)}
	\thanks{X. Luo is with  College of Computer Science and Software Engineer, Shenzhen University. (Email: xlluo@szu.edu.cn)}
	\thanks{Y. Chen and Z. He are with School of Computer Science and Technology, Harbin Institute of Technology, Shenzhen. (Email: YongyongChen.cn@gmail.com and zhenyuhe@hit.edu.cn)}
	\thanks{P. Wang is with SKL-IOTSC, Department of Computer and Information Science, University of Macau. (Email: pywang@um.edu.mo)}
	\thanks{Corresponding author: \textit{Pengyang Wang}.}
	}

\maketitle

\begin{abstract}
Computer-Aided Design (CAD), a fundamental technique in modern engineering, still relies heavily on manual workflows that are labor-intensive and time-consuming.
Recent deep learning approaches seek to automate CAD creation by representing a model as a sequence of discrete commands and parameters, and then generating them using autoregressive models or continuous diffusion operating in Euclidean embedding space.
However, continuous diffusion perturbs representations in a continuous Euclidean domain that does not reflect the inherently discrete and heterogeneous nature of CAD tokens, often producing perturbed representations that map to semantically invalid symbols.
To overcome this limitation, we propose a cascaded discrete diffusion framework for CAD generation, which consists of a command diffusion for generating CAD commands and a parameter diffusion conditioned on CAD commands.
Unlike isotropic Gaussian perturbation, the forward process of our approach operates directly over categorical token distributions using delicate transition matrices.
For commands, we adopt an absorbing-state transition matrix that progressively corrupts tokens to a designated symbol;
for parameters, we introduce specific transition matrices tailored to heterogeneous attributes: a Gaussian kernel for coordinate continuity, a scale-invariant kernel for dimensional values, and a prior-preserving kernel for boolean attributes.
The reverse process is achieved by two denoising networks: a Transformer-based encoder for command recovery, and a parameter network with extra local self-attention for command-level interaction and cross-attention for conditional injection.
Experiments on the DeepCAD dataset show that the proposed approach surpasses existing autoregressive and continuous diffusion models on unconditional generation metrics, while qualitative results validate effective controllability in conditional generation tasks.
Source codes will be released.
\end{abstract}


\section{Introduction}
\label{sec:intro}
\begin{figure}[t]
	\centering
	\begin{subfigure}{0.5\textwidth}
		\centering
		\includegraphics[width=\linewidth]{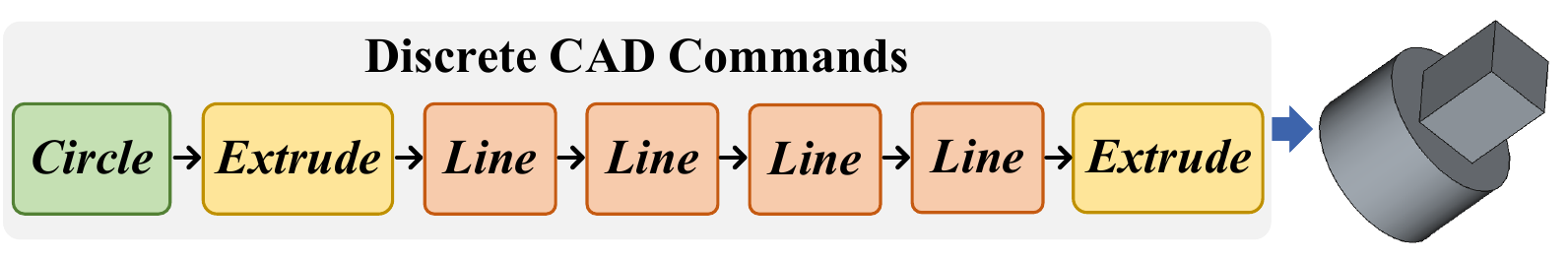}
		\vspace{-5mm}
		\caption{CAD construction}
		\hspace{-2mm}
		\label{fig:motivation_CAD}
	\end{subfigure}
	\begin{subfigure}{0.21\textwidth}
		\centering
		\includegraphics[width=\linewidth]{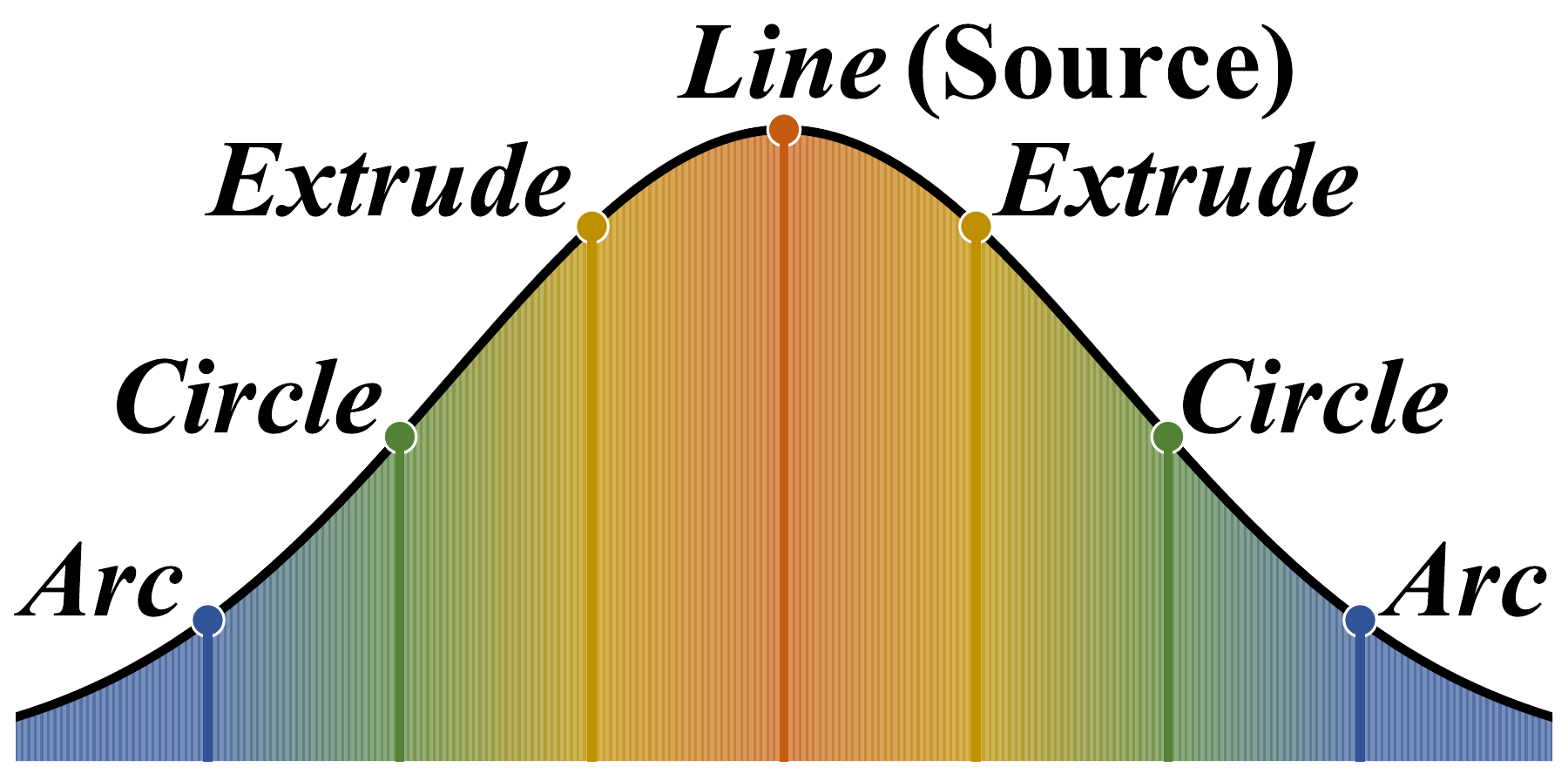}
		\vspace{-5mm}
		\caption{Continuous diffusion}
		\label{fig:motivation_continuous}
	\end{subfigure}
	\begin{subfigure}{0.27\textwidth}
		\centering
		\includegraphics[width=\linewidth]{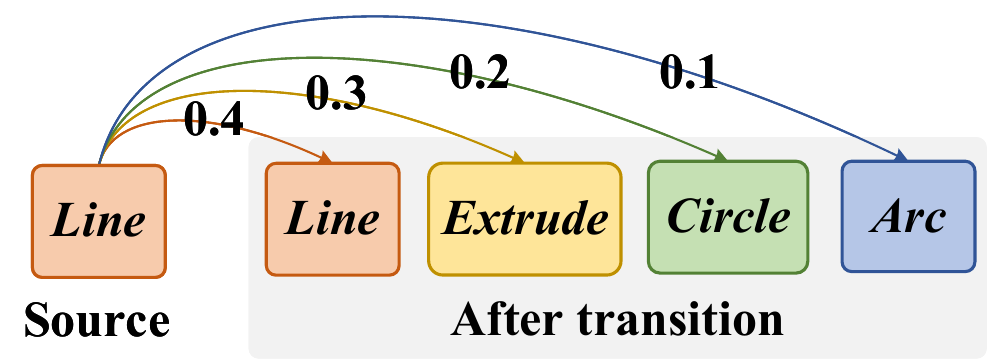}
		\vspace{-5mm}
		\caption{Discrete diffusion}
		\label{fig:motivation_discrete}
	\end{subfigure}
	\vspace{-5mm}
	\caption{
		(a) CAD construction is performed through a sequence of discrete commands (for clarity, parameters are omitted).
		(b) Existing continuous diffusion methods embed discrete commands into a continuous space and perturb them with isotropic Gaussian noise, which may result in invalid or semantically meaningless commands.
		(c) The proposed discrete diffusion operates directly over discrete token probabilities, perturbing only meaningful symbols and preserving semantic consistency.
	}
	\label{fig:motivation}
\end{figure}

Computer-Aided Design (CAD) serves as a cornerstone of modern engineering and manufacturing.
Conventional CAD workflows still rely heavily on manual operations, making them time-consuming and prone to human error.
Automating CAD design with machine learning models~\cite{Brepgen,CMT,HNC-CAD,PointerCAD,tii1,tii2,VQCAD} can substantially reduce design costs and improve reliability, offering great potential to accelerate industrial product development.
To enable CAD generation via deep learning, recent works~\cite{Deepcad,SkexGen,HNC-CAD,Diffusion-CAD,RECAD,CadVLM,tii3} represent a CAD model as a sequence of discrete commands and parameters, mimicking how designers create models step-by-step (see Fig.~\ref{fig:motivation_CAD}).
This representation, pioneered by DeepCAD~\cite{Deepcad}, organizes commands into two categories: sketch operations (e.g., \emph{Line}, \emph{Circle}) that define 2D geometric primitives, and extrusion operations (e.g., \emph{Extrude}) that construct 3D solids.
Each command is accompanied by a set of quantized parameters.
For example, a \emph{Circle} command is parameterized by its center coordinates $(x, y)$ and radius $r$.

Early methods rely on autoregressive models to generate CAD command sequences either directly in data space~\cite{Deepcad} or in a learned latent space~\cite{SkexGen,HNC-CAD}.
To improve sample diversity and controllability, Diffusion-CAD~\cite{Diffusion-CAD} introduced continuous diffusion into this domain by embedding discrete CAD tokens into a continuous vector space and perturbing them with isotropic Gaussian noise, as shown in Fig.~\ref{fig:motivation_continuous}.
However, this continuous embedding approach presents a fundamental limitation: CAD tokens are inherently discrete and lie in a non-Euclidean space, and their projection into a continuous domain fails to preserve semantic neighborhood relationships. 
Consequently, perturbed embeddings may correspond to invalid or semantically incoherent CAD commands, undermining the geometric validity of generated designs.
To overcome this limitation, we propose a cascaded discrete diffusion model (CDDM) for CAD generation.
Unlike continuous diffusion~\cite{LDM,DDPM,DDIM,ACD}, our approach operates directly over categorical token distributions and perturbs probabilities only among valid discrete states (see Fig.~\ref{fig:motivation_discrete}), thereby maintaining semantic consistency throughout the diffusion process.
Moreover, discrete diffusion~\cite{D3PM,VQ-diffusion,MD4,MDLM,SEDD,RADD,M2DM,M2D2M,LayoutDM,LayoutDiffusion} offers additional advantages for quantized CAD parameters.
Discrete transition matrices explicitly define structured corruption patterns, for instance, boolean attributes can be constrained to transition only among valid categories. 
In contrast, continuous diffusion applies isotropic noise in a continuous space, making such symbol-aware control infeasible for heterogeneous CAD parameters.

The proposed method adopts a cascaded architecture composed of two discrete diffusion modules: a command diffusion for generating CAD commands, followed by a parameter diffusion that predicts CAD parameters conditioned on the generated commands.
In the forward process, discrete diffusion corrupts categorical variables through a Markov chain parameterized by transition matrices, gradually converting them toward a steady distribution.
For the command diffusion, inspired by~\cite{Diffusionbert,VQ-diffusion}, we employ an absorbing-state transition matrix that progressively maps tokens to a designated absorbing symbol, effectively corrupting CAD commands.
For the parameter diffusion, CAD parameters are divided into three categories: coordinate, dimensional, and boolean parameters, each of which is assigned a tailored transition matrix.
Specifically, coordinate parameters use a Gaussian-based kernel to capture local spatial continuity; dimensional parameters adopt a scale-invariant transition scheme to enable diffusion over relative measurements; boolean parameters employ a prior-preserving kernel that restricts transitions to valid categories according to their empirical statistics.

In the reverse process, the model recovers valid commands and parameters from corrupted tokens via denoising networks.
The command diffusion network is implemented as a Transformer encoder~\cite{Transformer} equipped with multi-head self-attention layers.
The parameter diffusion network introduces three key modules: a global self-attention layer for long-range dependencies, a local self-attention layer that captures command-level parameter correlation, and a cross-attention layer for incorporating conditional information.
The local self-attention mechanism is the core innovation in this stage.
Since each command corresponds to multiple parameters, we flatten all parameters and apply a command-level attention mask that suppresses attention interactions between parameters belonging to different commands.

We evaluate the proposed method on the DeepCAD dataset~\cite{Deepcad}.
The experimental results show clear advantages of our cascaded discrete diffusion framework for CAD generation, which significantly outperforms existing autoregressive and continuous diffusion methods on unconditional generation.
We further demonstrate the controllability of our model on conditional generation tasks.
The main contributions of this paper are threefold:
\begin{itemize}
	\item We introduce the first cascaded discrete diffusion framework for CAD generation, which directly models categorical command and parameter tokens without embedding them into continuous space. By perturbing probability distributions only among valid discrete states, our method preserves semantic consistency and avoids invalid geometry commonly produced by continuous diffusion.
	\item We design symbol-aware transition mechanisms for heterogeneous CAD parameters, including a Gaussian kernel for coordinate continuity, a scale-invariant kernel for dimensional quantization, and a prior-preserving kernel for boolean attributes. These structured transition matrices enable controllable and token-valid corruption processes that are infeasible in continuous diffusion.
	\item We propose a command-aware parameter denoising network with local self-attention and cross-conditioning, which explicitly restricts parameter interactions within each command and supports conditional guidance.
\end{itemize}

\section{Related Works}
\label{sec:relatedworks}

In this section, we introduce recent studies on CAD generation.
A key challenge in applying machine learning to CAD model generation is to transform CAD data into representations suitable for computational processing.
Three major paradigms for CAD representation have been explored in the machine learning community: boundary representation (B-rep) ~\cite{Complexgen,Brepgen,CMT}, sequential representation~\cite{Deepcad,SkexGen,HNC-CAD,Diffusion-CAD,RECAD}, and textual representation~\cite{CAD-Llama,FlexCAD} for fine-tuning LLMs~\cite{LLaMA3,Gpt-4}.
For B-rep–based CAD generation, ComplexGen~\cite{Complexgen} designed chain complexes consisting of B-Rep geometric primitives of different orders and proposed a framework with a sparse CNN and a three-path Transformer decoder to model the B-Rep chain complex structure.
BrepGen~\cite{Brepgen} represented a B-rep model as a structured latent geometry organized in a hierarchical tree, and leveraged Transformer-based diffusion models to reconstruct the B-rep topology.
For sequential CAD generation, DeepCAD~\cite{Deepcad} was the first to represent CAD shapes as sequences of modeling operations, introducing an auto-regressive generative model for sequential modeling.
SkexGen~\cite{SkexGen} and HNC-CAD~\cite{HNC-CAD} further modeled CAD structures as hierarchical trees (entity–face–loop) and employed VQ-VAE~\cite{VQ-VAE} to learn discrete codebooks.
Diffusion-CAD~\cite{Diffusion-CAD} was the first to adopt diffusion models for sequential CAD generation, defining distinct embedding functions for discrete command types and continuous parameters to accommodate vectorized diffusion inputs.
Recently, several studies~\cite{CAD-Llama,FlexCAD} have explored the use of large language models (LLMs) for CAD generation.
These approaches constructed structured textual representations of CAD models and fine-tuned general-purpose LLMs such as LLaMA3~\cite{LLaMA3} and GPT-4~\cite{Gpt-4} to generate corresponding structured CAD text outputs.

\begin{figure*}[t]
	\centering
	\begin{center}
		\includegraphics[width=1.0 \textwidth]{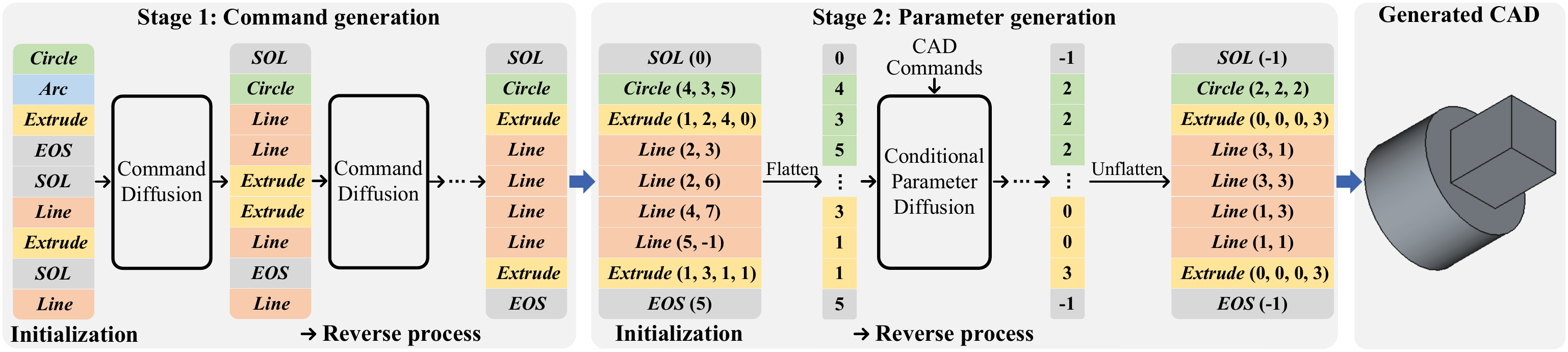}
	\end{center}
	\vspace{-0.4cm} 
	\caption{
		Pipeline of the proposed cascaded discrete diffusion model (CDDM) for CAD generation.
		The generation process consists of two sequential stages: the first stage produces discrete CAD commands through a command diffusion model, and the second stage reconstructs the corresponding CAD parameters using a conditional discrete diffusion model conditioned on the generated commands.
	}
	\label{fig:framework}
\end{figure*}

\section{Methods}
\label{sec:methods}

\subsection{Overview}
\label{sec:Overview}
By recent studies~\cite{Deepcad, Diffusion-CAD}, a CAD model $\mathcal{M}$ is represented as a sequence of discrete commands $\zeta$ and associated parameters $\theta$, as revealed in Table~\ref{table:cad}.
All continuous parameters are quantized into 256 discrete levels and encoded as 8-bit integers.
Consequently, the CAD generation task can be formulated as predicting a sequence of discrete CAD commands and their corresponding quantized parameters:
\begin{equation}
	p(\mathcal{M}) = p(\zeta, \theta),
\end{equation}
which can be further factorized into a command prior and a conditional parameter model:
\begin{equation}
	p(\mathcal{M}) = p(\zeta)\, p(\theta \mid \zeta).
\end{equation}
This factorization on discrete commands and parameters naturally motivates a cascaded diffusion architecture, where the first diffusion models command semantics, and the second diffusion learns geometric parameters conditioned on the generated commands.

In this paper, we propose a \textbf{cascaded discrete diffusion model (CDDM)}, a two-stage generative framework designed for CAD model synthesis.
As illustrated in Fig.~\ref{fig:framework}, the first stage employs a \emph{command diffusion model} to generate discrete CAD commands.
Conditioned on the generated commands, the second stage employs a \emph{parameter diffusion model} to infer the corresponding discrete parameters.
The rationale behind this cascaded design is that a single unified diffusion model tends to conflate structural semantics (commands) and geometric details (parameters), since both are represented as discrete tokens but exhibit distinct geometric semantics.
The forward processes for token corruption is introduced in Section~\ref{sec:Forward}, while the reverse processes for token recovery is presented in Section~\ref{sec:Reverse}.
Pipeline for conditional is introduced in Section~\ref{sec:condition}.

\subsection{Forward Process}
\label{sec:Forward}
A discrete data sample is represented as categorical variables $x_0 \in \{1, \dots, K\}^L$, where $K$ denotes the vocabulary size and $L$ is the sequence length. 
The forward process in discrete diffusion models gradually corrupts $x_0$ through a Markov chain of transition matrices $\{Q_t\}_{t=1}^T$.
The forward process can be expressed as:
\begin{equation}
	q(x_t \mid x_{t-1}) = \mathrm{Cat}(x_t; p = Q_t x_{t-1}),
	\label{eq:q_xt}
\end{equation}
\begin{equation}
	q(x_t \mid x_0) = \mathrm{Cat}(x_t; p = \bar{Q}_t x_0),
	\label{eq:q_x0}
\end{equation}
where $\bar{Q}_t = Q_1 Q_2 \cdots Q_t$ is the cumulative transition matrix, enabling direct sampling of $x_t$ from $x_0$.
A well-designed $Q_t$ corrupts data in a way that aligns with its native structure, thereby preserving underlying logical relationships.
In CAD generation, although both commands and parameters are represented as discrete tokens, they possess distinct structural and semantic dependencies. 
A uniform transition matrix~\cite{D3PM} corrupts tokens indiscriminately, undermining the geometric continuity and procedural logic essential to CAD.

\noindent \textbf{Transition matrices for command diffusion.}
For the command diffusion model, inspired by prior studies~\cite{Diffusionbert,VQ-diffusion}, 
we adopt an absorbing-state transition matrix $Q_t^{\text{absorb}} \in \mathbb{R}^{(K^{\langle \zeta \rangle}+1)\times(K^{\langle \zeta \rangle}+1)}$ to better capture the sequential and categorical nature of CAD commands:
\begin{equation}
	Q_t^{\text{absorb}} =
	\begin{bmatrix}
		\alpha_t + \beta_t & \beta_t & \beta_t & \cdots & 0 \\
		\beta_t & \alpha_t + \beta_t & \beta_t & \cdots & 0 \\
		\beta_t & \beta_t & \alpha_t + \beta_t & \cdots & 0 \\
		\vdots & \vdots & \vdots & \ddots & \vdots \\
		\gamma_t & \gamma_t & \gamma_t & \cdots & 1
	\end{bmatrix},
	\label{eq:Q_absorb}
\end{equation}
where $\gamma_t = 1 - \alpha_t - K\beta_t$ denotes the probability of transitioning to the absorbing state.
In this design, each command token evolves toward the absorbing state as diffusion proceeds, simulating the gradual erasure of its semantic content.

\begin{table}[ht]
	\caption{CAD commands and parameters. We categorize CAD parameters into three groups and design a dedicated transition matrix for each category: (1) Coordinate parameters (in \textcolor{red}{red}) represent the spatial positions of geometric entities; (2) Dimensional parameters (in \textcolor{green}{green}) describe geometric scales such as length, radius, or height; (3) Boolean parameters (in \textcolor{blue}{blue}) encode discrete design options, such as inner or outer operations in extrusion.
	We omit parameters for $\langle$\emph{SOL}$\rangle$ and $\langle$\emph{EOS}$\rangle$, as they are meaningless and would be filled with -1 in practical.
	The contents of this table come from DeepCAD.
	}
	\vspace{-0.15cm}
	\begin{tabular}{cl}
		\toprule
		Commands                     & Parameters                                                                                                                                                                                      \\ \midrule
		$\langle$\emph{SOL}$\rangle$ & $\Phi$                                                                                                                                                                                             \\ \midrule
		\emph{Line}                         & \textcolor{red}{$x, y$ : line end-point}                                                                                                                                                                           \\ \midrule
		\emph{Arc}                          & \begin{tabular}[c]{@{}l@{}}  \textcolor{red}{$x, y$ : arc end-point} \\ \textcolor{green}{$\alpha$ : sweep angle} \\ \textcolor{blue}{$f$ : counter-clockwise flag} \end{tabular}                                                                                    \\ \midrule
		\emph{Circle}                       & \begin{tabular}[c]{@{}l@{}}  \textcolor{red}{$x, y$ : center} \\ \textcolor{green}{$r$ : radius} \end{tabular}                                                                                                                               \\ \midrule
		\emph{Extrude}                      & \begin{tabular}[c]{@{}l@{}} \textcolor{red}{$\theta, \phi, \gamma$ : sketch plane orientation} \\ \textcolor{red}{$p_x, p_y, p_z$ : sketch plane origin} \\ \textcolor{green}{$s$ : scale of associated sketch profile} \\ \textcolor{green}{$e_1, e_2$ : extrude distances toward both sides} \\ \textcolor{blue}{$b$ : boolean type} \\ \textcolor{blue}{$u$ : extrude type} \end{tabular} \\ \midrule
		$\langle$\emph{EOS}$\rangle$ & $\Phi$                                                                                                                                                                                              \\ \bottomrule
	\end{tabular}
	\label{table:cad}
\end{table}

\noindent \textbf{Transition matrices for parameter diffusion.}
The CAD parameters associated with each command are highly heterogeneous, exhibiting distinct geometric semantics and statistical properties.
To this end, we design category-specific transition matrices tailored to different parameter type.
As summarized in Table~\ref{table:cad}, CAD parameters are categorized into three groups: (1) \emph{Coordinate parameters} represent the spatial positions of geometric entities; (2) \emph{Dimensional parameters} describe geometric scales such as length, radius, or height; and (3) \emph{Boolean parameters} encode discrete design options, such as inner or outer operations in extrusion.
We omit parameters for $\langle$\emph{SOL}$\rangle$ and $\langle$\emph{EOS}$\rangle$, as they are meaningless and would be filled with -1 in practical.

\begin{figure}[t]
	\centering
	\begin{subfigure}{0.19\textwidth} 
		\centering
		\includegraphics[width=\linewidth]{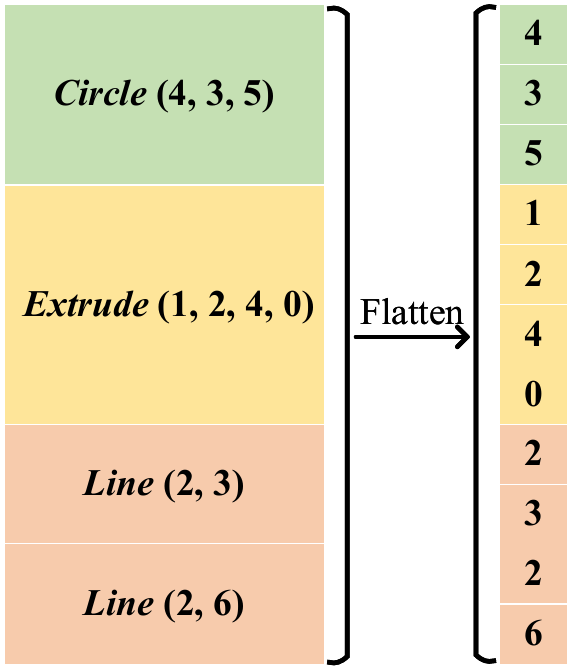}
		\vspace{-5mm}
		\caption{CAD parameters}
		\label{fig:attention1}
	\end{subfigure}
	\hspace{2mm}
	\begin{subfigure}{0.24\textwidth}
		\centering
		\includegraphics[width=\linewidth]{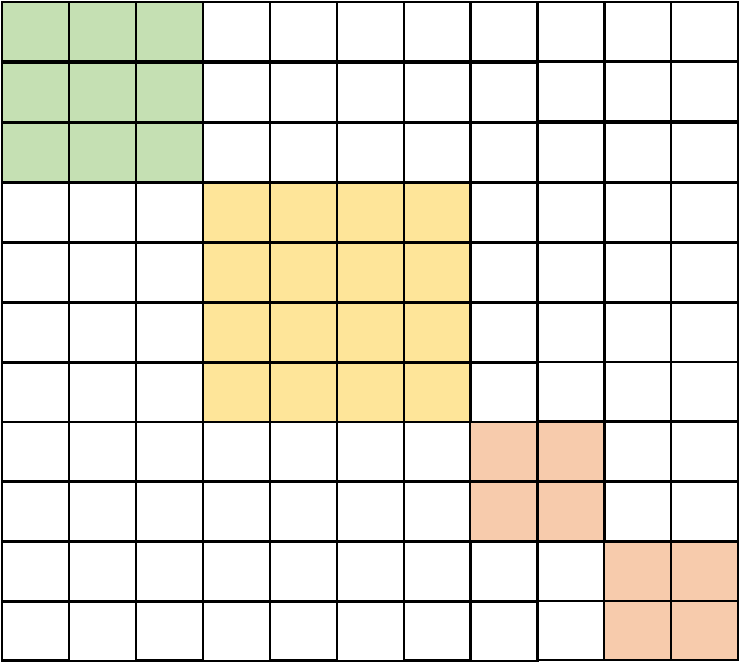}
		\vspace{-4mm}
		\caption{Local attention map}
		\label{fig:attention2}
	\end{subfigure}
	\vspace{-1mm}
	\caption{
		Schematic of the attention mask used for command-level local attention.
		(a) illustrates the flattened parameter sequence corresponding to different CAD commands.
		(b) shows the local attention mask, where colored cells indicate valid interactions (mask value $M_{ij}=0$) and blank cells indicate masked positions (mask value $M_{ij}=-\infty$).
	}
	\label{fig:attention}
\end{figure}

As the transition matrices for command diffusion, we employ an absorbing-state transition scheme for CAD parameter categories to progressively erase their semantic information. Specifically, all categories follow a unified absorbing-state formulation:
\begin{equation}
	Q_t^{\text{param}} =
	\begin{bmatrix}
		(1-\gamma_t)\,Q_t^{\text{base}} & \mathbf{0} \\
		\gamma_t \mathbf{1}^\top & 1
	\end{bmatrix},
	\label{eq:Q_param}
\end{equation}
where $Q_t^{\text{base}} \in \mathbb{R}^{K^{\langle \theta \rangle} \times K^{\langle \theta \rangle}}$ is the category-specific transition kernel and $\gamma_t$ controls the transition rate to the absorbing state.

\emph{Coordinate parameters.}
Although represented as discrete tokens, coordinate parameters correspond to continuous spatial quantities.
For example, the tokens 99 and 101 are semantically adjacent to 100.
To model this local continuity, we employ a Gaussian-based transition kernel:
\begin{equation}
	(Q_t^{\text{gaussian}})_{ij} = 
	(1-\alpha_t)\,
	\frac{
		\exp\!\left[-\frac{(i-j)^2}{2\sigma_t^2}\right]
	}{
		\sum\limits_{k=1}^{K} 
		\exp\!\left[-\frac{(k-j)^2}{2\sigma_t^2}\right]
	}
	+ \alpha_t \delta_{ij},
	\label{eq:Q_gauss_base}
\end{equation}
where $\sigma_t$ controls the local diffusion strength, 
$\alpha_t$ denotes the self-preservation probability, 
and $\delta_{ij}$ is the Kronecker delta, defined as 1 if $i=j$ and 0 if $i \neq j$.
This Gaussian design ensures that tokens are more likely to diffuse toward nearby coordinate values than toward distant ones.

\emph{Dimensional parameters.}
For parameters that describe geometric scales, a fixed absolute change can have drastically different meanings depending on magnitude (e.g., a 1\% relative increment from 100 to 101 versus a 100\% relative increment from 1 to 2).
To achieve scale invariance, we define a scale-invariant transition matrix $Q_t^{\text{scale}}$ where perturbations are measured relatively rather than absolutely:
\begin{equation}
	(Q_t^{\text{scale}})_{ij}
	=
	(1-\alpha_t)\,
	\frac{
		\exp\!\left[-\mu
		\left(\frac{i-j}{i+j}\right)^{\!2}\right]
	}{
		\sum\limits_{k=1}^{K}
		\exp\!\left[-\mu
		\left(\frac{k-j}{k+j}\right)^{\!2}\right]
	}
	\;+\;
	\alpha_t\,\delta_{ij},
	\label{eq:Q_scale_base}
\end{equation}
where $\mu$ controls the smoothness of the scale-invariant kernel.
This design ensures that diffusion is scale-invariant by measuring proximity through the relative measure $\frac{|i-j|}{i+j}$, not the absolute distance $|i-j|$.

\emph{Boolean parameters.}
For categorical parameters whose valid values lie within a fixed set $\mathcal{K}=\{1,\dots,k\}$, we employ a prior-preserving transition kernel restricted to this valid range:
\begin{equation}
	(Q_t^{\text{prior}})_{ij} =
	\begin{cases}
		\alpha_t + (1-\alpha_t)\tilde{p}_i, & \text{if } i,j \in \mathcal{K}, ~i=j, \\[2pt]
		(1-\alpha_t)\tilde{p}_i, & \text{if } i,j \in \mathcal{K},~i\neq j, \\[2pt]
		\tilde{p}_i, & \text{if } i \in \mathcal{K},~j \notin \mathcal{K}, \\[2pt]
		0, & \text{otherwise},
	\end{cases}
	\label{eq:Q_prior}
\end{equation}
where 
\[
\tilde{p}_i=\frac{\mathrm{count}(i)}{\sum_{j=1}^k \mathrm{count}(j)}
\]
is the empirical prior of category $i$ estimated from the training data.

This design yields two key properties.
First, it strictly prevents diffusion into invalid categories by enforcing zero probability outside $\mathcal{K}$.
Second, transitions among valid states are guided by a data-driven prior $\tilde{p}$:
a valid state $j\in\mathcal{K}$ remains unchanged with probability 
$\alpha_t + (1-\alpha_t)\tilde{p}_j$, and transitions to any other valid state $i\neq j$ with probability $(1-\alpha_t)\tilde{p}_i$.
If the previous state lies outside the valid range ($j\notin\mathcal{K}$), it is projected back to the valid domain according to the prior $\tilde{p}_i$.
Together, these constraints ensure that diffusion is both semantically valid and statistically consistent with observed category frequencies.

\begin{algorithm}[t]
	\caption{Training Process of CDDM}
	\LinesNumbered
	\DontPrintSemicolon
	\small
	\setstretch{1.2}
	\label{algo:training}
	\KwIn{Training data $\zeta_0 \sim q(\zeta_0)$ and $\theta_0 \sim q(\theta_0)$, transition matrices $\{Q_t^{\langle \zeta \rangle}\}_{t=1}^T$ and $\{Q_t^{\langle \theta \rangle}\}_{t=1}^T$, cumulative matrices $\{\bar{Q}_t^{\langle \zeta \rangle}\}_{t=1}^T$ and $\{\bar{Q}_t^{\langle \theta \rangle}\}_{t=1}^T$}
	\KwOut{Learned denoising models $z^{\langle \zeta \rangle}$ and $z^{\langle \theta \rangle}$}
	\For{each iteration}{
		Sample data $\zeta_0 \sim q(\zeta_0)$ and $\theta_0 \sim q(\theta_0)$\;
		Sample diffusion step $t \sim \mathrm{Uniform}(\{1,\dots,T\})$\;
		Sample $\zeta_t \sim q(\zeta_t \mid \zeta_0) = \mathrm{Cat}(p = \bar{Q}_t^{\langle \zeta \rangle} \zeta_0)$ and $\theta_t \sim q(\theta_t \mid \theta_0) = \mathrm{Cat}(p = \bar{Q}_t^{\langle \theta \rangle} \theta_0)$\;
		Compute $q(\zeta_{t-1} \mid \zeta_t, \zeta_0)$ and $q(\theta_{t-1} \mid \theta_t, \theta_0)$ using Eq.~\ref{eq:posterior}\;
		Predict $p_z(\zeta_{t-1} \mid \zeta_t)$ with $z^{\langle \zeta \rangle}$\;
		Predict $p_z(\theta_{t-1} \mid \theta_t, \zeta_0)$ with  $z^{\langle \theta \rangle}$\;
		Compute training losses using Eq.~\ref{eq:loss1} and Eq.~\ref{eq:loss2}\;
		Update $z^{\langle \zeta \rangle}$ and $z^{\langle \theta \rangle}$ by gradient descent\; 
	}
\end{algorithm}

\subsection{Reverse Process}
\label{sec:Reverse}
The reverse process of discrete diffusion progressively reconstructs samples consistent with the observed data, starting from the corrupted state $x_T$:
\begin{equation}
	p_z(x_{t-1} \mid x_t) = \sum_{\hat{x}_0} q(x_{t-1} \mid x_t, \hat{x}_0) \, p_z(\hat{x}_0 \mid x_t),
	\label{eq:reverse}
\end{equation}
where $q(x_{t-1} \mid x_t, x_0)$ is the tractable posterior derived from Bayes’ rule:
\begin{equation}
	q(x_{t-1} \mid x_t, x_0) \propto q(x_t \mid x_{t-1}) \, q(x_{t-1} \mid x_0).
	\label{eq:posterior}
\end{equation}
We aim to design a neural network that estimates the posterior distribution $p_z(\hat{x}_0 \mid x_t)$ for both command diffusion and parameter diffusion.
To this end, we adopt Transformer-based architectures $z^{\langle \zeta \rangle}$ and $z^{\langle \theta \rangle}$ for command generation and parameter generation.

\noindent \textbf{Denoising network for command diffusion.}
Given a corrupted command sequence $\zeta_t \in \mathbb{R}^l$ at time step $t$, the denoising network $z^{\langle \zeta \rangle}$ aims to recover the clean command sequence $\zeta_0 \in \mathbb{R}^l$.
The network consists of a Stylization block~\cite{GLIDE,Motiondiffuse} that fuses $\zeta_t$ with the time step $t$, followed by a Transformer encoder~\cite{Transformer} that models global dependencies among command tokens.
Specifically, the Stylization block can be written as follows:
\begin{equation}
	\begin{split}
		(f^1_t, f^2_t) &= \phi_t(t), \\
		f^{\langle \zeta \rangle} = \mathrm{LN}(\phi_{\zeta}(\zeta_t&)) \cdot (1 + f^1_t) + f^2_t,
	\end{split}
\end{equation}
where $\phi_t$ and $\phi_{\zeta}$ are linear projection layers, and $\mathrm{LN}$ denotes LayerNorm.
Next, the stylized features $f^{\langle \zeta \rangle}_t \in \mathbb{R}^{l \times d}$ are then processed by a Transformer encoder:
\begin{equation}
	\begin{split}
		Q = f^{\langle \zeta \rangle} W^Q, \ K &= f^{\langle \zeta \rangle} W^K, \ V = f^{\langle \zeta \rangle} W^V, \\
		\text{Attention}(Q, K, &V) = \text{softmax}\left(\frac{QK^T}{\sqrt{d_k}}\right)V.
	\end{split}
	\label{eq:attention}
\end{equation}
For brevity, we omit the positional encoding, multi-head attention, residual connection, and other standard Transformer components.
The output of the encoder is taken as the predicted clean command sequence $\hat{\zeta}_0$.

\noindent \textbf{Denoising network for parameter diffusion.}
Since each command is associated with multiple parameters, we first flatten all parameters across commands into a single sequence, and repeat each command token to match the number of parameters it corresponds.
For example, given commands \emph{Circle}, \emph{Line}, \emph{Line} with parameter sets (4, 3, 5), (2, 3), and (2, 6), respectively, we construct the parameter sequence $[4, 3, 5, 2, 3, 2, 6]$ and the command sequence $[\emph{Circle}, \emph{Circle}, \emph{Circle}, \emph{Line}, \emph{Line}, \emph{Line}, \emph{Line}]$.

\begin{algorithm}[t]
	\LinesNumbered
	\DontPrintSemicolon
	\small
	\setstretch{1.2}
	\caption{Sampling Process of CDDM}
	\label{algo:sampling}
	\KwIn{Transition matrices $\{Q_t^{\langle \zeta \rangle}\}_{t=1}^T$ and $\{Q_t^{\langle \theta \rangle}\}_{t=1}^T$, learned reverse model $z^{\langle \zeta \rangle}$ and $z^{\langle \theta \rangle}$}
	\KwOut{Generated CAD sample $\langle \zeta_0, \theta_0 \rangle$}
	
	Initialize with absorbing states $\zeta_T = (K^{\langle \zeta \rangle}+1)^L$\;
	\For{$t=T$ \KwTo $1$}{
		Compute model prediction $p_z(\zeta_{t-1} \mid \zeta_t)$ with $z^{\langle \zeta \rangle}$\;
		Sample $\zeta_{t-1} \sim p_z(\zeta_{t-1} \mid \zeta_t)$ using Eq.~\ref{eq:reverse}\;
	}
	Initialize with absorbing states $\theta_T = (K^{\langle \theta \rangle}+1)^L$\;
	\For{$t=T$ \KwTo $1$}{
		Compute model prediction $p_z(\theta_{t-1} \mid \theta_t, \zeta_0)$ with $z^{\langle \theta \rangle}$\;
		Sample $\theta_{t-1} \sim p_z(\theta_{t-1} \mid \theta_t, \zeta_0)$ using Eq.~\ref{eq:reverse}\;
	}
	\Return $\langle \zeta_0, \theta_0 \rangle$\;
\end{algorithm}

We aim to design a denoising network $z^{\langle \theta \rangle}$ that predicts the clean parameter sequence $\theta_0 \in \mathbb{R}^L$ from the corrupted parameters $\theta_t \in \mathbb{R}^L$ at time step $t$, conditioned on the repeated command sequence $\zeta_t \in \mathbb{R}^L$.
The overall architecture of $z^{\langle \theta \rangle}$ follows the same backbone as $z^{\langle \zeta \rangle}$, consisting of a Stylization block followed by a Transformer encoder.
The Transformer encoder in $z^{\langle \theta \rangle}$ is more structured: each block contains a global self-attention layer, a local self-attention layer to model command-level correlations, and a cross-attention layer that incorporates conditional command information.

The global attention layer follows Eq.~(\ref{eq:attention}) with $f^{\langle \theta \rangle}$ (from the Stylization block) as input.
To model correlations only within the parameters belonging to the same command, we construct a command-level attention mask $M \in \mathbb{R}^{L \times L}$ (Fig.~\ref{fig:attention}), where entries corresponding to different commands are suppressed.
The local self-attention operation is formulated as:
\begin{equation}
	\begin{split}
		Q = f^{\langle \theta \rangle}_1 W^Q, \ K &= f^{\langle \theta \rangle}_1 W^K, \ V = f^{\langle \theta \rangle}_1 W^V, \\
		\text{LocalAttention}(Q, K&, V) = \text{softmax}\left(\frac{QK^T}{\sqrt{d_k}} + M\right)V,
	\end{split}
	\label{eq:local_attention}
\end{equation}
where $f^{\langle\theta\rangle}_1$ denotes the output of the global attention, and the mask is defined as
$$
M_{ij} = \begin{cases}
	0 & \text{if} \ \ \zeta_i=\zeta_j, \\
	-\infty & \text{if} \ \ \zeta_i \neq \zeta_j.
\end{cases} \\
$$
Finally, the cross-attention layer takes the output $f^{\langle \theta \rangle}_2$ of the local self-attention as queries, and the command feature $f^{\langle \zeta \rangle}$ as keys and values, enabling explicit interaction between parameter tokens and their corresponding command embeddings:
\begin{equation}
	\begin{split}
		Q = f^{\langle \theta \rangle}_2 W^Q, K &= f^{\langle \zeta \rangle} W^K, V = f^{\langle \zeta \rangle} W^V, \\
		\text{CrossAttention}(Q, &K, V) = \text{softmax}\left(\frac{QK^T}{\sqrt{d_k}}\right)V.
	\end{split}
	\label{eq:cross_attention}
\end{equation}

\noindent \textbf{Training objectives.}
The command diffusion model uses the same loss formulation as standard discrete diffusion:
\begin{equation}
	\mathcal{L} = \mathbb{E}_{q(\zeta_{0:T})} \Bigg[ \sum_{t=1}^T D_{\mathrm{KL}} \big( q(\zeta_{t-1} \mid \zeta_t, \zeta_0) \,\|\, p_z(\zeta_{t-1} \mid \zeta_t) \big) \Bigg].
	\label{eq:loss1}
\end{equation}
For the parameter diffusion model, the loss introduces an additional conditioning term on the command sequence $\zeta_0$:
\begin{equation}
	\mathcal{L} = \mathbb{E}_{q(\theta_{0:T})} \Bigg[ \sum_{t=1}^T D_{\mathrm{KL}} \big( q(\theta_{t-1} \mid \theta_t, \theta_0) \,\|\, p_z(\theta_{t-1} \mid \theta_t, \zeta_0) \big) \Bigg].
	\label{eq:loss2}
\end{equation}
The overall training and sampling procedures for the proposed CDDM are summarized in Algorithm~\ref{algo:training} and Algorithm~\ref{algo:sampling}, respectively.

\begin{table*}[ht]
	\centering
	\small
	\caption{Comparison with state-of-the-art methods. Best results are highlighted in \textbf{bold}.}
	\vspace{-2mm}
	\begin{tabular}{clc|cccccc}
		\toprule
		Representation            & Method        & Venue  & COV$\uparrow$ & MMD$\downarrow$ & JSD$\downarrow$ & Novelty$\uparrow$ & Unique$\uparrow$ & Invalidity$\downarrow$ \\ \midrule
		\multirow{2}{*}{B-rep}    
		& BrepGen~\cite{Brepgen}       & TOG24  & 78.16    & 1.02    & \textbf{0.09}    & 99.9        & 97.6       & 20.2           \\
		& CMT~\cite{CMT}           & ICCV25 & 75.71    & \textbf{0.92}    & 1.02    & 99.0        & \textbf{99.8}       & 29.9           \\ \midrule
		\multirow{2}{*}{Text}     & CAD-Llama~\cite{CAD-Llama}     & CVPR25 & 65.60    & 1.19    & 0.82    & 92.1        & 92.6       & 6.6           \\
		& FlexCAD~\cite{FlexCAD}       & ICLR25  & 80.46    & 0.96    & 0.66        & 97.1       & -     & -         \\ \midrule
		\multirow{4}{*}{Sequence} & DeepCAD~\cite{Deepcad}       & ICCV21 & 73.58    & 1.73    & 1.45    & 86.3        & 91.4       & 10.2           \\
		& SkexGen~\cite{SkexGen}       & ICML22 & 79.42    & 1.60    & 1.13    & 98.6        & 97.3       & 22.4           \\
		& HNC-CAD~\cite{HNC-CAD}       & ICML23 & 80.41    & 1.47     & 0.98    & 96.3        & 96.1       & 11.7           \\
		& Diffusion-CAD~\cite{Diffusion-CAD} & TVCG25 & 79.11    & 1.62    & 1.07    & 99.5        & 97.1       & 4.1           \\ \midrule
		Sequence                  & CDDM (ours)   & -      & \textbf{89.10}    & 0.96    & 1.07    & \textbf{100}        & 95.4       & \textbf{4.0}           \\ \bottomrule
	\end{tabular}
	\label{table:comp_sota}
\end{table*}

\subsection{Conditional Generation}
\label{sec:condition}
To support conditional CAD generation, CDDM in Fig.~\ref{fig:framework} can be naturally extended from the unconditional to the conditional setting.
Given an input condition $c$, we first encode it into a latent representation $f^{\langle c \rangle}$.
For the command diffusion, we introduce an additional cross-attention layer following the self-attention in Eq.~(\ref{eq:attention}), where the queries come from the command features, while both keys and values are derived from the conditional features.
For the parameter diffusion, we modify the cross-attention in Eq.~(\ref{eq:cross_attention}) by replacing its keys and values with the fused feature $f^{\langle \zeta \rangle}+f^{\langle c \rangle}$, allowing parameters to be guided jointly by command and conditional information.

In this work, we investigate two types of conditions: command length and point clouds.
For command length, we encode the one-hot length indicator using a linear layer and apply it only to the command diffusion stage.
For CAD generation from point clouds (i.e., reverse engineering~\cite{SIGNet,StepbyStep,TransCAD,CAD-Recode}), we extract point cloud features using PointNet~\cite{PointNet} and use them as conditional inputs for both command and parameter diffusion.

\section{Experiments}
\label{sec:experiments}

\subsection{Dataset and Experimental Settings}
\noindent \textbf{Dataset.}
To evaluate the effectiveness of the proposed CDDM, we conduct experiments on the DeepCAD dataset~\cite{Deepcad}, a large-scale collection of parametric CAD designs containing over 178,238 models, split into 90\%/5\%/5\% for training/validation/testing.
As summarized in Table~\ref{table:cad}, each CAD sample is represented as a sequence of commands and their associated parameters. The dataset contains six command categories: $\langle$\emph{SOL}$\rangle$, \emph{Line}, \emph{Arc}, \emph{Circle}, \emph{Extrude}, and $\langle$\emph{EOS}$\rangle$, leading to a command vocabulary size of $K^{\langle \zeta \rangle}=6$ for the command diffusion.
All effective parameters are uniformly quantized into 256 discrete values, while the parameters associated with $\langle$\emph{SOL}$\rangle$ and $\langle$\emph{EOS}$\rangle$ are assigned a fixed value of $-1$. This results in a parameter vocabulary size of $K^{\langle \theta \rangle}=257$ for the parameter diffusion.
The maximum sequence lengths are set to 60 for command diffusion and 280 for parameter diffusion.

\noindent \textbf{Training settings.}
We jointly optimize both diffusion models.
The denoising network $z^{\langle \zeta \rangle}$ for command diffusion uses 8 Transformer blocks, while $z^{\langle \theta \rangle}$ for parameter diffusion employs 4 Transformer blocks.
All hidden dimensions are set to 256, and the total number of diffusion steps $T$ is fixed to 100.
We adopt the schedule from LayoutDiffusion~\cite{LayoutDiffusion} to sample the transition coefficients $\alpha_t$ and $\gamma_t$, which changes token categories in the late phase of the forward process.
The variance of the Gaussian kernel ($\sigma_t^2$ in Eq.~(\ref{eq:Q_gauss_base})) is set to 2.0, while the smoothness parameter of the scale-invariant kernel ($\mu$ in Eq.~(\ref{eq:Q_scale_base})) is set to 1.0.
Both diffusion models are trained for 100 epochs using the Adam optimizer~\cite{Adam} with a learning rate of $4 \times 10^{-5}$. Training is performed on a single NVIDIA RTX 4090 GPU, requiring approximately 15 minutes per epoch.

\begin{figure*}[ht]
	\centering
	\begin{center}
		\includegraphics[width=0.98 \textwidth]{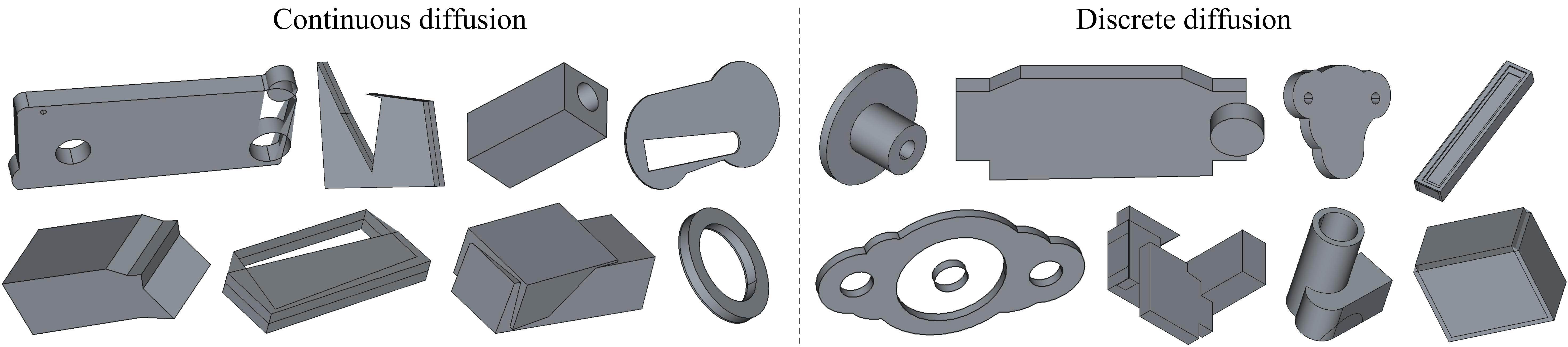}
	\end{center}
	\vspace{-0.3cm} 
	\caption{
		Visualization of CAD models generated by continuous diffusion and discrete diffusion.
	}
	\label{fig:vis_cont_disc}
\end{figure*}

\begin{figure*}[ht]
	\centering
	\begin{center}
		\includegraphics[width=0.95 \textwidth]{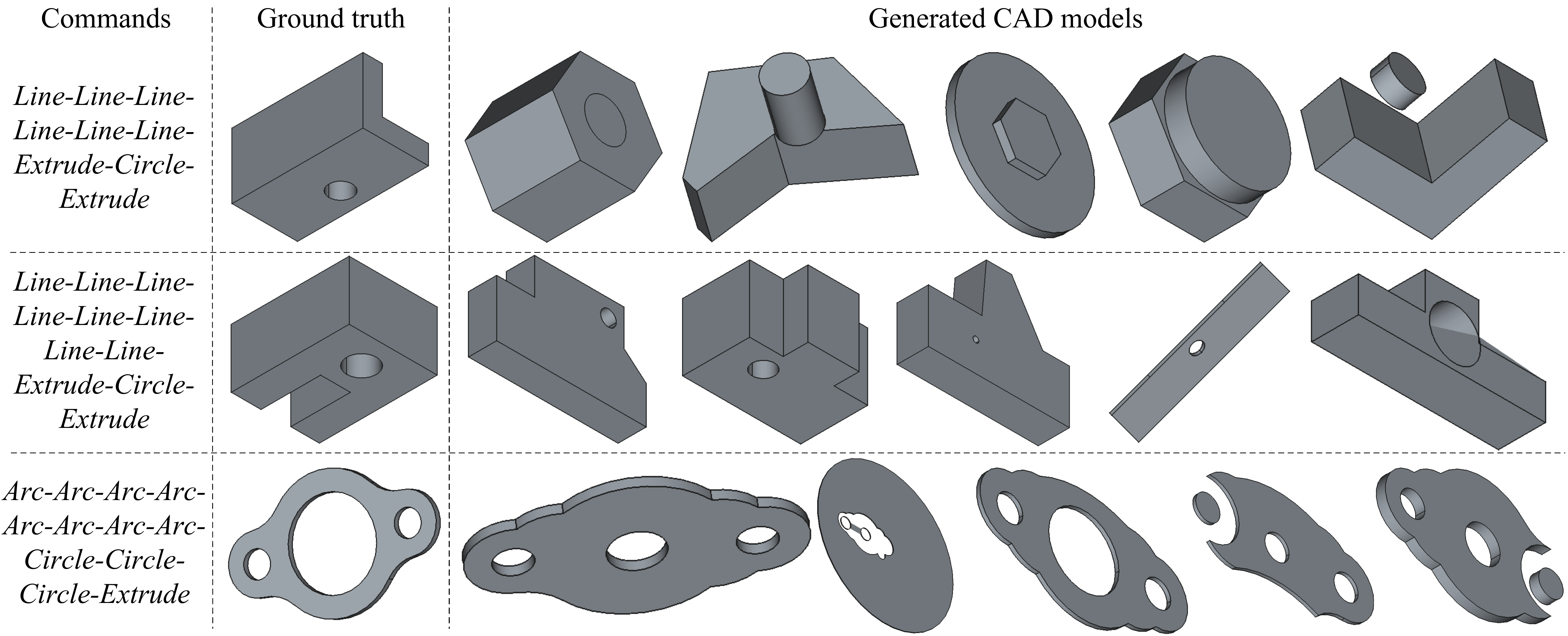}
	\end{center}
	\vspace{-0.3cm} 
	\caption{
		Visualization of CAD models using ground truth commands.
	}
	\label{fig:vis_gt_commands}
\end{figure*}

\noindent \textbf{Testing settings.}
For a fair comparison, we follow the evaluation protocols used in the continuous diffusion baseline~\cite{Diffusion-CAD}.
For unconditional generation, we report six evaluation metrics: \textbf{Coverage (COV)}, \textbf{Minimum Matching Distance (MMD)}, \textbf{Jensen–Shannon Divergence (JSD)}, \textbf{Novelty}, \textbf{Unique}, and \textbf{Invalidity}.
\textbf{COV} measures sample diversity by computing the proportion of generated shapes that are sufficiently close to real shapes under Chamfer Distance (CD).
\textbf{MMD} evaluates fidelity by calculating the minimum chamfer-based distance between each generated point cloud and the closest ground-truth sample.
\textbf{JSD} quantifies the distributional discrepancy between generated and real point sets.
\textbf{Novelty} represents the fraction of generated designs absent from the training set.
\textbf{Unique} measures the proportion of samples that appear exactly once in the generated set.
\textbf{Invalidity} reports the percentage of command sequences that fail to produce valid CAD models.
For conditional generation, we provide qualitative evaluations by visualizing synthesized CAD models.

\subsection{Unconditional Generation}

\subsubsection{Comparison with State-of-the-Art Methods}
We compare the proposed CDDM with three categories of CAD generative models: 
(1) methods based on B-rep representation, (2) methods using sequence-based representation, and (3) methods using text-based representation. 
The results are reported in Table~\ref{table:comp_sota}. 
As shown in the table, CDDM achieves new state-of-the-art performance on COV, Novelty, and Invalidity. 
In particular, CDDM surpasses the continuous diffusion model Diffusion-CAD~\cite{Diffusion-CAD} by a significant margin, demonstrating the clear advantage of discrete diffusion in modeling inherently discrete CAD structures.

\subsubsection{Verification of Model Architecture}
In this subsection, we investigate three questions:  
(1) Is the proposed cascaded architecture more effective than a unified diffusion architecture?  
(2) Is discrete diffusion superior to continuous diffusion for CAD generation?  
(3) Between the command and parameter diffusion modules, which imposes the primary performance bottleneck?

\noindent \textbf{Cascaded diffusion vs.\ unified diffusion.}  
We first compare CDDM with a unified architecture.
In the unified setting, each command token is concatenated with its corresponding parameters, and all commands are then concatenated into a single 1D sequence:
\[
(\zeta_1,\theta_{11},\theta_{12},\ldots,\theta_{1j},\zeta_2,\theta_{21},\ldots,\theta_{2j},\ldots,\zeta_n,\theta_{n1},\ldots,\theta_{nj}),
\]
where $\zeta_i$ denotes the command at step $i$ and $\theta_{ij}$ denotes its parameters.  
In CDDM, command tokens range from 0 to 5 and parameter tokens from 0 to 255.  
To distinguish tokens in the unified setting, we shift parameter values by 255, mapping them into the range 256--271.

Experiments show that the unified diffusion model exhibits an Invalidity rate exceeding 90\%, indicating that it rarely generates valid CAD sequences.
In contrast, the cascaded architecture achieves an Invalidity rate below 5\%.  
The unified model incorrectly treats command and parameter tokens as homogeneous symbols, mixing structural semantics with geometric details and ultimately leading to a low validity.

\begin{figure*}[ht]
	\centering
	\begin{center}
		\includegraphics[width=0.9 \textwidth]{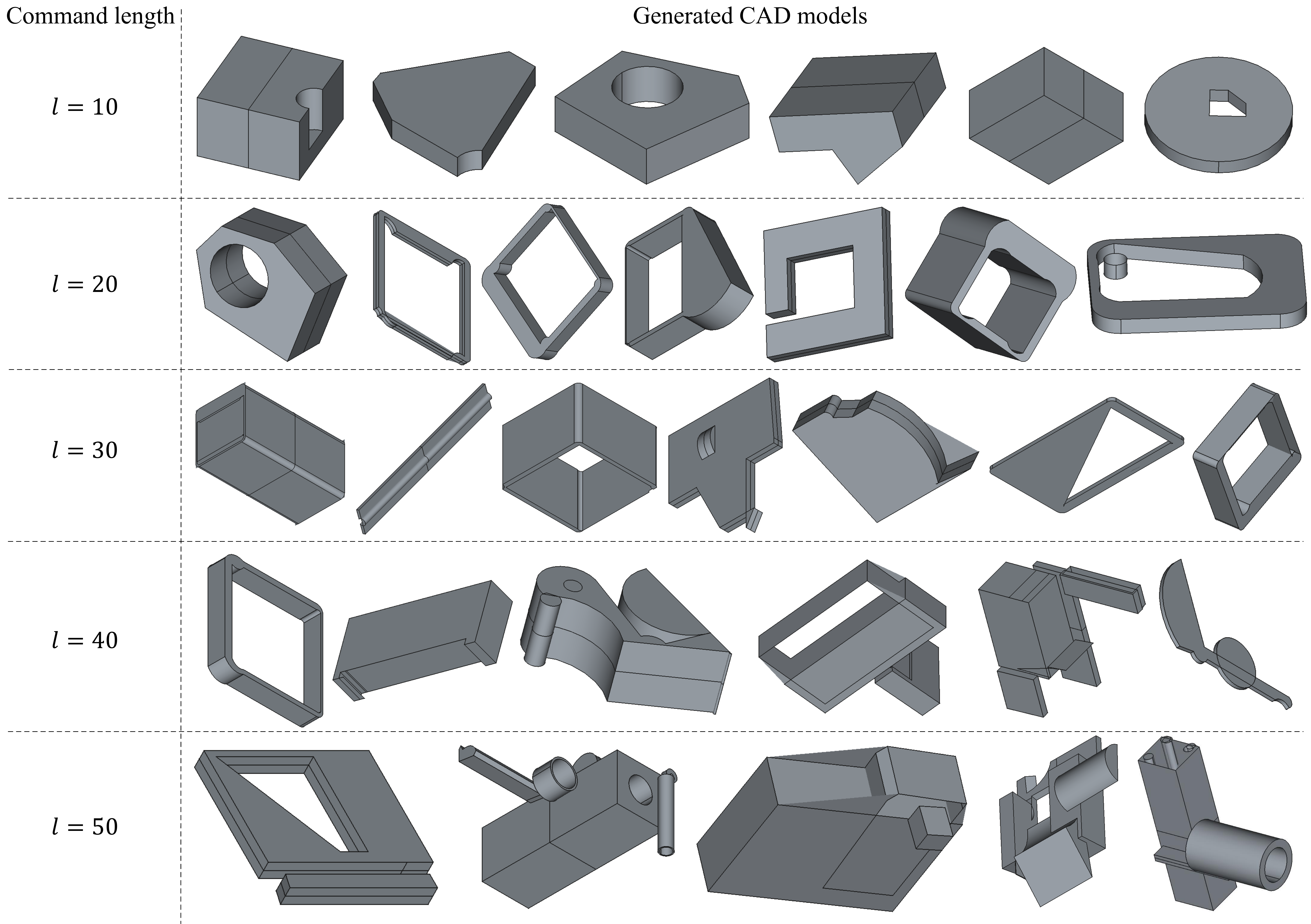}
	\end{center}
	\vspace{-0.3cm} 
	\caption{
		Visualization of generated CAD models under the condition of command length.
	}
	\label{fig:vis_command_len}
\end{figure*}

\noindent \textbf{Discrete diffusion vs.\ continuous diffusion.}  
We evaluate discrete and continuous diffusion for the task of parameter generation.
We do not re-implement continuous diffusion for both commands and parameters, as Diffusion-CAD~\cite{Diffusion-CAD} has extensively explored this setting.
To adapt continuous diffusion, each quantized parameter is normalized to $[0,1]$ by dividing by 255.

Table~\ref{table:comp_dis_con} reports the comparison.
Discrete diffusion outperforms continuous diffusion across all metrics.
Visual comparisons in Fig.~\ref{fig:vis_cont_disc} further show that continuous diffusion often fails to produce complex CAD structures.

\begin{table}[t]
	\centering
	\small
	\caption{Comparison of discrete diffusion and continuous diffusion.}
	\vspace{-2mm}
	\begin{tabular}{l|cccccc}
		\toprule
		Diffusion  & COV$\uparrow$ & MMD$\downarrow$ & JSD$\downarrow$ & Nov$\uparrow$ & Uni$\uparrow$ & Inval$\downarrow$ \\ \midrule
		Continuous   & 82.20    & 1.17    & 2.93    & \textbf{100}        & \textbf{100}       & 5.11           \\
		Discrete     & \textbf{89.10}    & \textbf{0.96}    & \textbf{1.07}    & \textbf{100}        & 95.36       & \textbf{4.00}           \\ \bottomrule
	\end{tabular}
	\label{table:comp_dis_con}
\end{table}

\begin{table}[t]
	\centering
	\small
	\caption{Exploration of upper-bound performance of CDDM.}
	\vspace{-2mm}
	\begin{tabular}{l|cccccc}
		\toprule
		Mode  & COV & MMD & JSD & Nov & Uni & Inval \\ \midrule
		Upper-bound   & \textbf{90.60}    & \textbf{0.91}    & \textbf{0.97}    & \textbf{100}        & \textbf{97.02}       & \textbf{2.21}           \\
		CDDM     & 89.10    & 0.96    & 1.07    & \textbf{100}        & 95.36       & 4.00           \\ \bottomrule
	\end{tabular}
	\label{table:expl_upper}
\end{table}

\noindent \textbf{Upper-bound exploration.}  
Parameter diffusion can be conditioned on either generated commands or ground-truth commands.
To explore the upper-bound performance, we condition parameter diffusion on ground-truth command sequences.
As shown in Table~\ref{table:expl_upper}, using ground-truth commands yields slightly better results, revealing that parameter diffusion, rather than command diffusion, is the main performance bottleneck.  
Figure~\ref{fig:vis_gt_commands} illustrates that our parameter diffusion module generates diverse and geometrically plausible CAD models when conditioned on given commands.

\subsubsection{Verification of Model Components}
We next evaluate the effectiveness of major model components, including the symbol-specific transition matrices and the attention mechanisms in the parameter denoising network. We finally analyze the effect of different diffusion steps.

\begin{table}[t]
	\centering
	\scriptsize 
	\caption{Exploration on transition matrices.}
	\vspace{-2mm}
	\begin{tabular}{ccc|cccccc}
		\toprule
		\multicolumn{3}{c|}{Parameter type} & \multirow{2}{*}{COV} & \multirow{2}{*}{MMD} & \multirow{2}{*}{JSD} & \multirow{2}{*}{Nov} & \multirow{2}{*}{Uni} & \multirow{2}{*}{Inval} \\ \cmidrule{1-3}
		Coord      & Dimen      & Bool      &                      &                      &                      &                      &                      &                       \\ \midrule
		$Q^{\text{unif}}$          &  $Q^{\text{unif}}$           &   $Q^{\text{unif}}$         & 79.40                     & 1.17                     & 3.74                     & \textbf{100}                     & 98.96                     & 7.85                      \\ 
		$Q^{\text{unif}}$    &  $Q^{\text{scale}}$          &  $Q^{\text{prior}}$        & 86.30                     & 1.01                     & 1.62                     & \textbf{100}                     & \textbf{99.10}                     & 4.32                      \\
		$Q^{\text{gauss}}$    &  $Q^{\text{unif}}$          &  $Q^{\text{prior}}$        & 86.80                     & 1.00                     & 1.32                     & \textbf{100}                     & 97.57                     & 4.26                      \\
		$Q^{\text{gauss}}$    &  $Q^{\text{scale}}$          &  $Q^{\text{unif}}$        & 87.50                     & 1.05                     & \textbf{1.03}                     & \textbf{100}                     & 94.87                     & 6.25                      \\ \midrule
		$Q^{\text{gauss}}$    &  $Q^{\text{scale}}$          &  $Q^{\text{prior}}$          & \textbf{89.10}    & \textbf{0.96}    & 1.07                 & \textbf{100}                     & 95.36                     & \textbf{4.00}                      \\ \bottomrule
	\end{tabular}
	\label{table:expl_mat}
\end{table}

\begin{figure*}[ht]
	\centering
	\begin{center}
		\includegraphics[width=0.96 \textwidth]{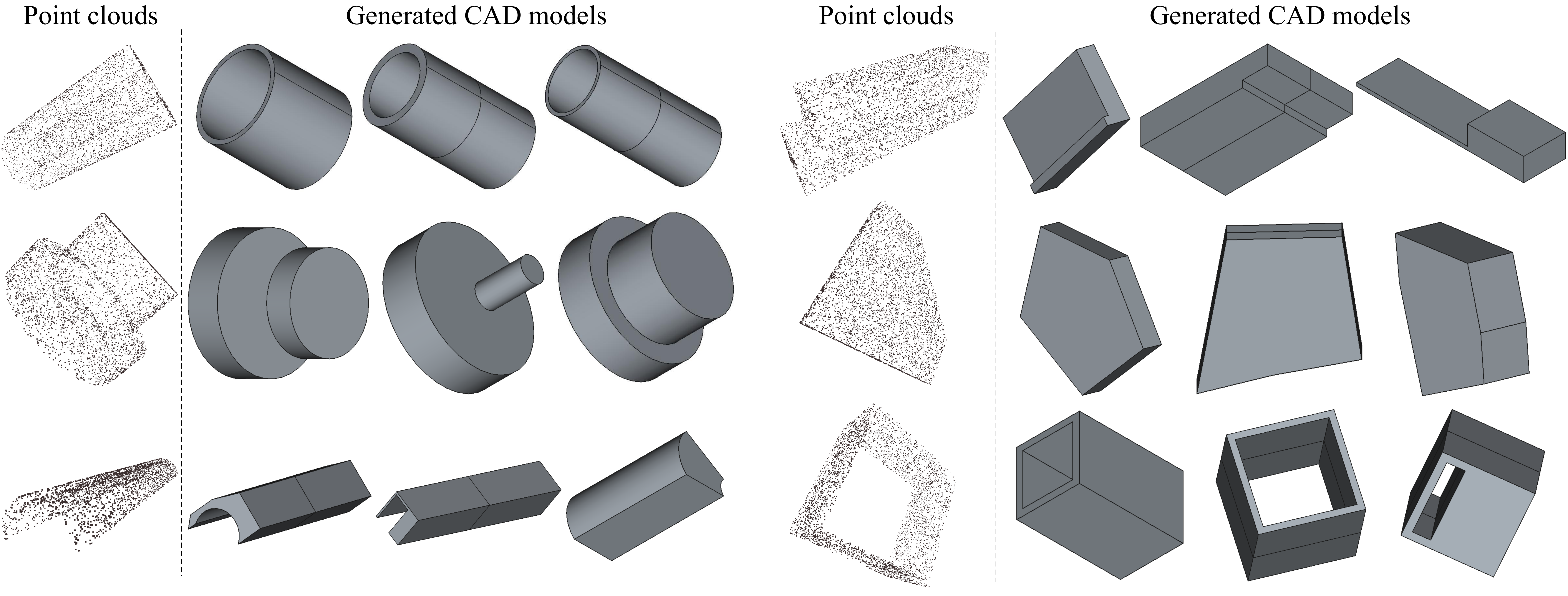}
	\end{center}
	\vspace{-0.3cm} 
	\caption{
		Visualization of generated CAD models under the condition of point clouds.
	}
	\label{fig:vis_pcs}
\end{figure*}

\noindent \textbf{Exploration of transition matrices.}  
For command diffusion, we use a uniform absorbing transition matrix $Q^{\text{unif}}$.
For parameter diffusion, we employ mixed transition matrices tailored to different parameter types:  
Gaussian ($Q^{\text{gauss}}$) for coordinate parameters, scale-invariant ($Q^{\text{scale}}$) for dimensional parameters, and prior-preserving ($Q^{\text{prior}}$) for boolean parameters.  
To evaluate their necessity, we replace each of $Q^{\text{gauss}}$, $Q^{\text{scale}}$, and $Q^{\text{prior}}$ with $Q^{\text{unif}}$.
If we use $Q^{\text{unif}}$ to corrupt all parameters, the transition matrix for the parameter diffusion degenerated to transition matrix for command diffusion. 

The results are reported in Table~\ref{table:expl_mat}.
If we use $Q^{\text{unif}}$ to corrupt all parameters, the transition matrix for the parameter diffusion degenerated to transition matrix for command diffusion.
We find that using $Q^{\text{unif}}$ for boolean parameters degrades the performance of Invalidity, and using $Q^{\text{unif}}$ for coordinate and dimensional parameters would degrades the performance of COV, MMD, and JSD.
We can conclude that the scheme of mixed transition matrices achieves a dominant results compared to other schemes.

\noindent \textbf{Impact of attention layers.}  
The parameter denoising network includes a local self-attention layer for modeling correlations within each command and a global self-attention layer for long-range dependencies.  
In Table~\ref{table:verif_attention}, we verify the impact of these two layers: we compare the performance of using a single attention layer and using both two layers.
From this table, using only the global attention is superior than using only the local attention, while using two attention mechanisms outperforms using single attention.

\noindent \textbf{Exploration of diffusion steps.}  
Continuous diffusion typically uses 1000 steps, which may be unnecessary for discrete diffusion.  
We test time-step settings of 100, 200, 500, and 1000 (Table~\ref{table:expl_steps}).  
Performance does not show a clear preference for any diffusion step.  
Since fewer steps also provide faster sampling, we use 100 steps as a practical trade-off.

\begin{table}[t]
	\centering
	\small
	\caption{Effectiveness verification of global and local attention layers.}
	\vspace{-2mm}
	\begin{tabular}{cc|cccccc}
		\toprule
		\multicolumn{2}{c|}{Attention} & \multirow{2}{*}{COV} & \multirow{2}{*}{MMD} & \multirow{2}{*}{JSD} & \multirow{2}{*}{Nov} & \multirow{2}{*}{Uni} & \multirow{2}{*}{Inval} \\ \cmidrule{1-2}
		Global         & Local         &                      &                      &                      &                          &                         &                             \\ \midrule
		$\checkmark$   &               & 87.50                     & 0.99                     & 1.14                     & \textbf{100}                         & 95.48                        & \textbf{3.02}                            \\
		&  $\checkmark$             & 76.50                     & 1.20                     & 2.68                     & \textbf{100}                         & \textbf{98.22}                        & 4.82                            \\
		$\checkmark$  & $\checkmark$             & \textbf{89.10}    & \textbf{0.96}    & \textbf{1.07}    & \textbf{100}        & 95.36       & 4.00                            \\ \bottomrule
	\end{tabular}
	\label{table:verif_attention}
\end{table}

\begin{table}[t]
	\centering
	\small
	\caption{Exploration on diffusion steps.}
	\vspace{-2mm}
	\begin{tabular}{l|cccccc}
		\toprule
		Diffusion steps  & COV & MMD & JSD & Nov & Uni & Inval \\ \midrule
		100   & \textbf{89.10}    & \textbf{0.96}    & 1.07    & \textbf{100}        & 95.36       & 4.00           \\
		200   & 86.60    & 1.06    & 1.01    & \textbf{100}        & 94.67       & 3.81           \\ 
		500   & 88.20    & 1.08    & 1.51    & \textbf{100}        & 95.21       & \textbf{2.89}           \\ 
		1000   & 87.70    & 1.04    & \textbf{0.70}    & \textbf{100}        & \textbf{96.69}       & 3.57           \\ \bottomrule
	\end{tabular}
	\label{table:expl_steps}
\end{table}

\subsection{Conditional Generation}
This section evaluates the controllability of our method through qualitative experiments on two conditions: command length and point clouds.
Fig.~\ref{fig:vis_command_len} illustrates generated CAD models conditioned on varying command lengths (10, 20, 30, 40, and 50).
As expected, the complexity of the generated structures increases consistently with the length constraint.
Fig.~\ref{fig:vis_pcs} presents results conditioned on point cloud inputs, where 4000 points are sampled per CAD model.
As can be seen,  the model successfully reconstructs diverse and geometrically plausible CAD sequences from the point clouds.
These experiments validate the effectiveness of the proposed CDDM in handling different conditional generation scenarios.

\section{Conclusions}
\label{sec:conclusions}
In this paper, we proposed a cascaded discrete diffusion framework for Computer-Aided Design (CAD) generation, factorizing this task into a command diffusion followed by a parameter diffusion conditioned on the predicted commands.
Unlike prior autoregressive and continuous diffusion models, our method operated directly on categorical command and parameter tokens, thereby avoiding the semantic inconsistencies introduced by perturbing continuous embeddings.
In the forward process, we designed a set of symbol-aware transition matrices for heterogeneous CAD commands and parameters: 
commands were corrupted through an absorbing-state transition matrix, while coordinate, dimensional, and boolean parameters were diffused using Gaussian-based, scale-invariant, and prior-preserving kernels, respectively.
In the reverse process, we employed a Transformer-based command denoiser and a parameter network incorporating global attention, command-level local attention, and cross-conditioning to reconstruct valid CAD sequences.
Experimental evaluations on the DeepCAD dataset demonstrated that the proposed framework achieved superior performance over existing autoregressive and continuous diffusion baselines in unconditional generation tasks.

\bibliographystyle{IEEEtran}
\bibliography{sample-base}

\end{document}